\documentclass[10pt,twocolumn,letterpaper]{article}

\usepackage{iccv}
\usepackage{times}
\usepackage{epsfig}
\usepackage{graphicx}
\usepackage{amsmath}
\usepackage{amssymb}
\usepackage{bm}
\usepackage{xspace}
\usepackage{gensymb}
\usepackage{capt-of}
\usepackage{enumitem}
\usepackage[dvipsnames]{xcolor}
\usepackage{multirow}
\usepackage{flushend}
\usepackage[accsupp]{axessibility} 
\usepackage[numbers,sort]{natbib}

\newcommand{\oursname}{VariTex\xspace}

\definecolor{gray}{gray}{0.4}

\newcommand{\reffig}[1]{Fig.~\ref{#1}}

\newcommand{\myparagraph}[1]{\vskip 0.5em\noindent\textbf{#1.}}

\usepackage[pagebackref=true,breaklinks=true,letterpaper=true,colorlinks,bookmarks=false]{hyperref}

\iccvfinalcopy

\ificcvfinal\fi

\begin{document}

\title{\oursname: Variational Neural Face Textures}

\author{Marcel C. B\"uhler$^{1}$ \quad Abhimitra Meka$^{2}$ \quad Gengyan Li$^{1,2}$ \quad Thabo Beeler$^{2}$ \quad Otmar Hilliges$^{1}$\vspace{0.1cm} \\
$^1$ETH Zurich \quad $^2$Google\\
{\small \href{https://mcbuehler.github.io/VariTex}{https://mcbuehler.github.io/VariTex}}
} \vspace{1cm}

\twocolumn[{
\renewcommand\twocolumn[1][]{#1}%
\maketitle
\begin{center}
  \newcommand{\teaserwidth}{\textwidth}
  \vspace{-0.7cm}
  \centerline{\includegraphics[width=\textwidth]{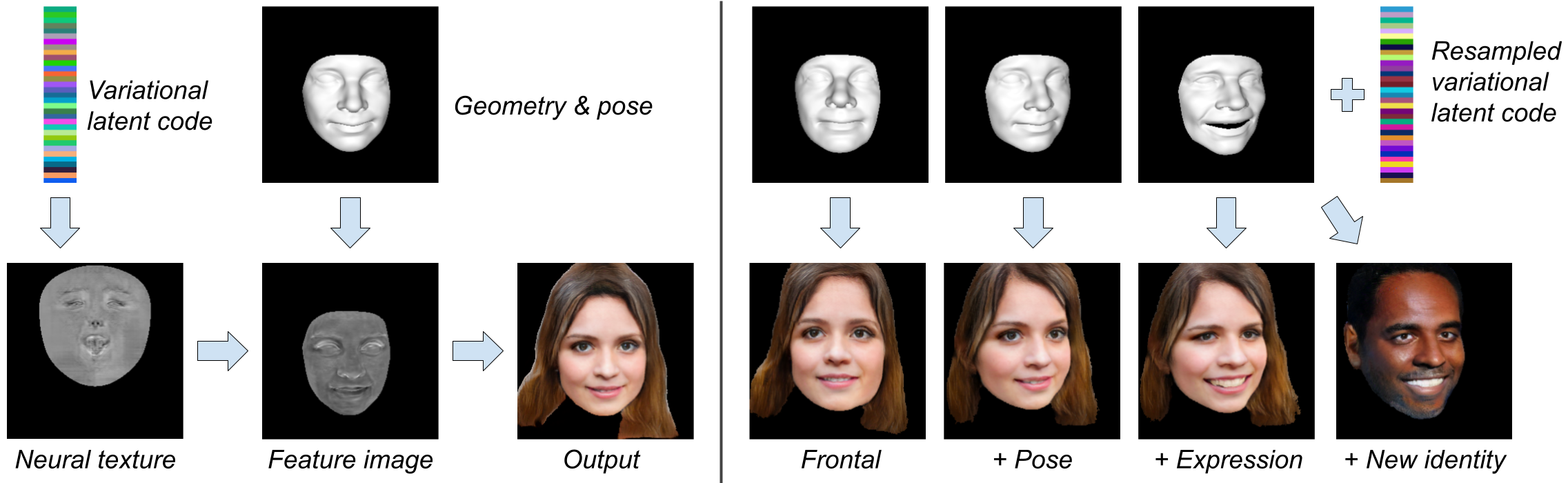}}
    \captionof{figure}{
    \oursname{} generalizes person-specific neural textures to variational textures. This allows to control both pose and expressions via explicit 3D geometries (\reffig{fig:qualitative}) and to sample novel identities (\reffig{fig:sampling}). Our method generates images under fine head pose and expression control, while maintaining geometric consistency over a large range of these parameters (\reffig{fig:comparison_pose} and Tbl.~\ref{tbl:consistency}).}
    \label{fig:teaser}
\end{center}
}]

\ificcvfinal\thispagestyle{empty}\fi

\begin{abstract}
Deep generative models can synthesize photorealistic images of human faces with novel identities.
However, a key challenge to the wide applicability of such techniques is to provide independent control over semantically meaningful parameters: appearance, head pose, face shape, and facial expressions.
In this paper, we propose \oursname{} - to the best of our knowledge the first method that learns a variational latent feature space of neural face textures, which allows sampling of novel identities.
We combine this generative model with a parametric face model and gain explicit control over head pose and facial expressions.
To generate complete images of human heads, we propose an additive decoder that adds plausible details such as hair.
A novel training scheme enforces a pose-independent latent space and in consequence, allows learning a one-to-many mapping between latent codes and pose-conditioned exterior regions.  
The resulting method can generate geometrically consistent images of novel identities under fine-grained control over head pose, face shape, and facial expressions. This facilitates a broad range of downstream tasks, like sampling novel identities, changing the head pose, expression transfer, and more.
\end{abstract}

\section{Introduction}
The ability to generate images with user-controlled parameters, such as identity-specific appearance, pose, and expressions would have many applications in computer graphics and vision. 
Synthesizing photorealistic images of novel human faces has recently been made possible through deep generative adversarial networks~\cite{goodfellow2014generative,karras2019style,karras2020analyzing} or variational auto encoders~\cite{kingma2014vae}, that learn the distribution of real faces to generate new identities. 
However, such methods typically do not provide semantic control over shape, pose, and facial expressions. This results in undesired global appearance changes across different generated images, for example, a change in identity when viewing from a different angle.

In order to gain more control over the generated images, recent work conditions neural networks on explicit 3D geometries~\cite{kim2018deep,tewari2020stylerig,tewari2020pie,thies2020neural,KowalskiECCV2020config,deng2020disentangleddiscofacegan,GIF2020}. Promising results have been shown by first generating the 2D face image from a learned latent space, and then attempting to rig it using graphics techniques in a geometrically consistent manner~\cite{tewari2020stylerig,tewari2020pie}. 
This approach suffers from an inherent disadvantage: since the image synthesis is performed in 2D space only, it is hard to enforce consistency under 3D manipulation. 
Strong supervision via multi-view images or multi-pose data from monocular videos at \textit{training} time can alleviate this to some degree. 
However, the inherent 2D nature of the solution prohibits a truly 3D consistent solution for novel \textit{test} poses, views, and expressions. This problem particularly manifests itself when synthesizing poses and expressions that lie outside the distribution of the training data (Fig.~\ref{fig:comparison_pose}).
To increase geometric faithfulness, recent work has attempted to learn the distribution of faces in 3D, for example by leveraging neural textures to represent 3D scenes in texture space ~\cite{thies2019deferred,thies2020neural,raj2020anr}. 
Especially when trained from video, rendering 3D geometry with \emph{neural} textures has been shown to produce highly consistent outputs for multiple poses and expressions, albeit at the cost of having to learn a texture \textit{per subject}.

Our work generalizes subject-specific neural textures~\cite{thies2019deferred,thies2020neural,grigorev2021stylepeople} to \emph{variational} neural textures, enabling geometry-aware synthesis of \emph{novel} identities (see \reffig{fig:teaser} and~\ref{fig:sampling}). Neural textures represent the appearance of a 3D surface as 2D feature maps. 
In contrast to prior works, which are trained per subject, \emph{variational} neural textures do not require strong supervision in the form of multi-view images or minutes-long video sequences as input \emph{for each identity}.  
Instead, they are generated by sampling from an underlying latent distribution of neural face textures. 
Importantly, this latent space is learned in a \textit{self-supervised} scheme from \emph{monocular} RGB images \textit{without requiring any annotations}. 
We use a parametric face model \cite{blanz1999morphable} in combination with a differentiable renderer to provide fine-grained control over face shape, head pose, and facial expression. 
This is sufficient to generate face interiors that preserve a subject's identity across pose and expression, but does not model other important details, such as ears, hair, and the mouth interior.
To attain complete images of human heads, we propose a pose-aware additive decoder that generates features for visually plausible details (e.g., facial hair). We devise a novel training regime that allows the additive decoder to learn a one-to-many mapping and in consequence to generate the exterior face region conditioned on different head poses from the same latent code (see \reffig{fig:teaser}).

We propose \oursname{}: Variational Neural Face Textures -- a method to sample novel identities and synthesize consistent faces in multiple poses and expressions (Fig. \ref{fig:teaser} and~\ref{fig:qualitative}).

We demonstrate state-of-the-art (SoA) photo-realistic results for geometric control (Fig.~\ref{fig:qualitative}), novel identity image synthesis (Fig.~\ref{fig:sampling}), and novel pose synthesis (Fig.~\ref{fig:teaser} and~\ref{fig:comparison_pose}). Our method achieves higher visual identity-consistency than related work (Fig.~\ref{fig:comparison_pose}). Quantitatively, we compare embedding distances between frontal and posed faces via a SoA face recognition network~\cite{deng2019arcface} (Tbl.~\ref{tbl:consistency}). Finally, we conduct a user study (Sec.~\ref{sec:results_userstudy}), where participants rate consistency for posed faces and overall photo-realism. 

In summary, we make the following contributions:
\begin{enumerate}[noitemsep]
\item \oursname, the first method for learning a variational latent feature space for neural face textures - allowing to sample novel identities.

\item Combining the generative power of learned facial textures with the explicit control of a parametric face model enables fine-grained control over facial expressions, head pose, face shape, and appearance.

\item We synthesize plausible outputs for difficult regions where no 3D geometries are available (e.g., hair, ears, and the mouth interior).

\item We show that our method is more identity consistent under geometric transformations.

\end{enumerate}

\section{Related Work}

We briefly review related work on image synthesis of human faces, particularly those that leverage differentiable rendering and neural textures. 

\begin{table}[ht!]
\small 
\begin{center}
\begin{tabular}{lcc}
\hline
\multirow{2}{*}{\textbf{Method}} & \textbf{Pose-independent} & \textbf{Sampling}\\
&\textbf{texture}&\textbf{novel identities}\\
\hline
UV-GAN~\cite{deng2018uv}    & RGB   & $\times$      \\
DNR~\cite{thies2019deferred} & neural  & $\times$      \\
NVP~\cite{thies2020neural}    & neural   & $\times$    \\
\hline
ConfigNet~\cite{KowalskiECCV2020config} & $\times$ & $\checkmark$ \\
GIF~\cite{GIF2020} & $\times$ & $\checkmark$ \\
DiscoFaceGAN~\cite{deng2020disentangleddiscofacegan} &$ \times$  & $\checkmark$  \\
\hline
\textbf{Ours} & neural & $\checkmark$ \\
\end{tabular}
\end{center}
\caption{Overview of most closely related methods. Texture-rendering based methods are not designed to sample new identities~\cite{thies2019deferred,thies2020neural,deng2018uv}. More generic synthesis methods~\cite{KowalskiECCV2020config,GIF2020,deng2020disentangleddiscofacegan} suffer from inconsistency under large pose variations (\reffig{fig:comparison_pose} and  Tbl.~\ref{tbl:consistency}) because they do not provide texture-level control over the face region. We propose a framework based on variational neural textures that can do both.
\label{tbl:relatedwork}}
\end{table}

\myparagraph{High-quality Face Synthesis}
Most modern methods for synthesizing natural images leverage generative adversarial networks (GAN)~\cite{goodfellow2014generative} or variational auto-encoders~\cite{kingma2014vae}. These methods have achieved a high level of photorealism~\cite{karras2019style,karras2020analyzing,park2019semanticspade,zhu2020sean,choi2018stargan,choi2020stargan,romero2019smit,buhler2020deepsee}.
Typically such methods learn to map from a low dimensional latent space to the distribution of 2D face images using convolutional neural networks. 
However, these latent spaces often entangle appearance and geometry~\cite{karras2019style,karras2020analyzing}, making novel pose or expression synthesis extremely difficult. 
Recent works started disentangling the latent space and adding more and more control~\cite{KowalskiECCV2020config,GIF2020,deng2020disentangleddiscofacegan,shoshan2021gan,tewari2020stylerig,tewari2020pie,garbin2020high,harkonen2020ganspace,sofgan}, for example, by learning disentanglement via statistical face models~\cite{blanz1999morphable} as strong priors~\cite{GIF2020,deng2020disentangleddiscofacegan}. 
Neural radiance fields~\cite{mildenhall2020nerf} have shown to render faces of very high quality~\cite{gafni2021dynamic,raj2021pixel,guo2021ad}. However, their generative variants~\cite{Schwarz2020NEURIPS,chan2021pi} still lack control over expressions.
In summary, generative modeling of photorealistic faces with artistic control remains a difficult challenge.
 
\myparagraph{Differentiable Rendering}
An alternative way to disentangle appearance from geometry and pose \emph{by design} is learning appearance in a UV space~\cite{deng2018uv,thies2019deferred,thies2020neural,raj2020anr,Meka:2020,shamai2019synthesizing,grigorev2021stylepeople}. 
Traditional computer graphics rendering pipelines require highly detailed 3D geometries, which are very expensive to obtain. Recently, Thies et al. proposed \emph{deferred neural rendering}~\cite{thies2019deferred}. Deferred neural rendering showed how deep neural networks can compensate imperfect 3D geometries and render highly photo-realistic imagery. 
Key components of these methods are \emph{neural textures}. 
Instead of using traditional textures in a pre-defined \emph{color space}, they leveraged the power of \emph{neural features} as a description of texture. 
As a difference to related works based on neural textures~\cite{thies2019deferred,raj2020anr}, our model is fully generative and allows sampling of new identities. Thies et al.~\cite{thies2019deferred} train person-specific models and Raj et al.~\cite{raj2020anr} optimize person-specific textures from videos. Our model is trained from monocular images alone.

\myparagraph{Neural Textures for Faces}
Previous methods using neural textures~\cite{thies2019deferred,thies2020neural} learn a \emph{person-specific} texture from multiview images or videos. Given enough data of a target person, they enable realistic animation of the facial expressions seen during training. 
The high expression fidelity and image quality comes at the cost of tightly coupling neural textures and rendering, which requires training a network per person. Furthermore, training neural textures per scene requires multiple views or minute-long sequences of the target person. An interesting challenge in the evolution of neural textures is to generalize them to single images and to novel identities. 
To this end, we frame the problem as a \emph{variational neural texture generation} task, followed by a texture-to-image translation task. This generalizes the texture and image generator to unseen identities, gives fine-grained control over head pose and facial expression and the generated images remain consistent under the manipulation of these parameters.

\section{\oursname: Variational Neural Textures}
\begin{figure*}[!ht]
    \centering
  \includegraphics[width=\textwidth]
                  {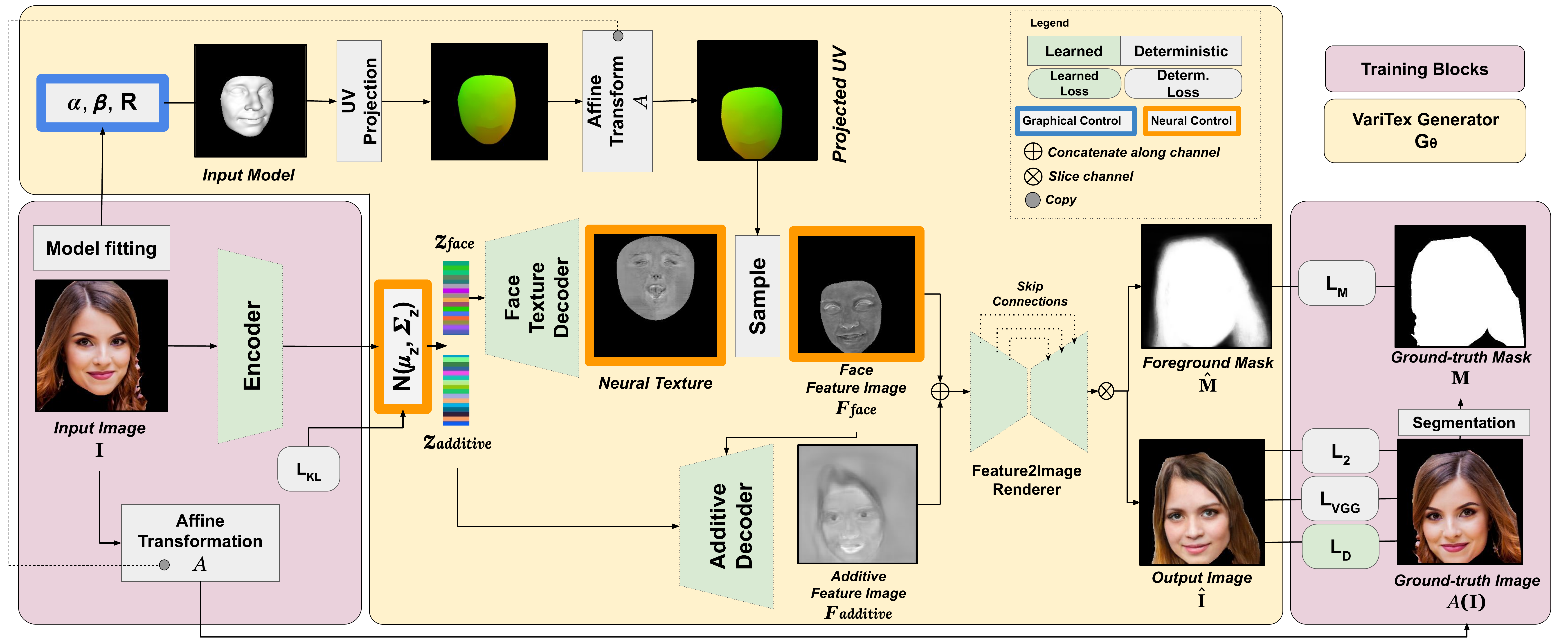}
    \caption{The objective of our pipeline is to learn a generator $\bm G_{\theta}$ that can synthesize face images with arbitrary novel identities whose expressions and pose can be controlled using face model parameters $\bm \alpha, \bm \beta$, and $\bm R$ (\reffig{fig:qualitative}). During training, we use unlabeled monocular RGB images ($ \bm I$) to learn a smooth latent space $\mathcal{N}(\bm \mu_z, \bm \Sigma_z)$ of natural face appearance using a variational encoder. A latent code $\bm z$ sampled from this space is then decoded to a novel face image. At test time, we draw samples to generate novel face images (\reffig{fig:sampling}). Our variationally generated \emph{neural textures} can also be stylistically interpolated to generate intermediate identities (supplementary material).
    \label{fig:pipeline}}
\end{figure*}

\subsection{Overview}\label{sec:method_overview}
We tackle the problem of controlled novel identity synthesis for faces, with the goal to disentangle appearance from pose and expression. To do so, we generalize person-specific neural textures~\cite{thies2019deferred,thies2020neural} to \emph{variational} neural textures by learning a distribution over identities that can map to a neural texture space. This allows generating an infinite number of neural textures that can be mapped on face geometries with arbitrary poses and expressions.

At the core of our method is a neural texture decoder that is trained in a self-supervised manner via neural rendering. The decoder learns to generate neural textures that follow a predefined layout given by the UV parameterization of a 3D morphable face model~\cite{gerig2018morphablebfm2017}, followed by a projection into image space and rendering as an RGB image.

 Intuitively, our network can render extreme poses despite being trained on largely frontal imagery, because the neural texture projection provides spatially aligned features. 
 Such neural rendering networks have been shown to generalize to poses unseen during training~\cite{thies2019deferred}.

\subsection{Problem Statement}
\label{sec:problem}
Our goal is to learn a generator $G_\theta$ that produces face images $\bm {\hat I}$ and foreground masks $\bm {\hat M}$
from a latent description for identity $\bm z \in \mathbb{R}^{d_z}$ and control signals for shape $\bm \alpha \in \mathbb{R}^{d_\alpha}$, expression  $\bm \beta \in \mathbb{R}^{d_\beta}$ and head pose $\bm R \in SO(3)$. Given a code for an identity $\bm z$ and a corresponding shape $\bm \alpha$, the generator should synthesize consistent images that preserve facial identity across different expressions $\bm \beta$ and poses $\bm R$. Equation~\ref{eq:problem} summarizes our problem statement:
\begin{align}
\label{eq:problem}
    (\bm {\hat I}, \bm {\hat M} ) = G_\theta(\bm z, \bm \alpha, \bm\beta, \bm R).
\end{align}
The distribution over $\bm z$ is learned from a large collection of monocular face images. 
Shape $\bm \alpha$ and expression $\bm \beta$ are the coefficients of a PCA-based 3D morphable face model~\cite{blanz1999morphable}, learned from a collection of 3D scans~\cite{gerig2018morphablebfm2017}. The pose $\bm R$ is a 3D rotation matrix. 
    
While we also train an image-to-latent-space encoder, we emphasize that this is more of a side effect. Our primary goal is to learn latent space from which \emph{novel} identities can be sampled and rendered under geometric control, as opposed to generating novel views of existing identities.

\subsection{Architecture Overview}\label{sec:architecture_overview}

Fig.~\ref{fig:pipeline} summarizes our method. During training, we use monocular RGB images to learn the underlying space of face appearance. This is done in the variational auto-encoder (VAE) framework~\cite{kingma2014vae}, where an encoder learns to map input face images to parameters of a normal distribution. These parameters can then be sampled to generate a latent code which is interpreted by the \oursname{} generator $\bm G_{\theta}$. We describe our training scheme in Sec. \ref{sec:training}. 

Unlike a traditional decoder of a VAE~\cite{kingma2014vae}, The \oursname{} Generator synthesizes face images in a geometry-aware manner. We use a parametric face model with consistent topology to map the 3D geometry of any face to a 2D texture layout. This 2D texture space serves as the domain over which feature maps for novel identities can be generated. The face model is then used to re-project the generated \emph{neural textures} from this layout to the output image space under any desired pose and expression. We describe this process in greater detail in Sec. \ref{sec:generator}.

The texture layout can only handle regions of the face geometry that are present in the face model. We use an additional network---the \emph{additive decoder}---to generate features for the exterior regions, such as hair, ears, and the mouth interior. 

Finally, a neural renderer converts the neural features into an RGB image and a plausible \emph{foreground mask}. The full generation process is described in detail in Sec.~\ref{sec:generator}.

\subsection{\oursname{} Generator}
\label{sec:generator}
This section describes the components of the~\oursname{} generator.
The generator consists of two decoders and a Feature2Image rendering network. The decoders produce a neural description of the desired output---the \emph{neural feature image}. The Feature2Image network turns these features into an RGB image and a corresponding foreground mask.

The generator allows a) to generate new identities by sampling latent codes $\bm z$ and shape coefficients $\bm \alpha$, and b) to manipulate expression $\bm \beta$ and pose $\bm R$.

The latent code for identity $\bm z \in \mathbb{R}^{256}$ can be sampled from a learned distribution $\mathcal{N}(\bm \mu_z, \bm \Sigma_z)$ or extracted from a reference image. It is split into two halves: $\bm z_{face} \in \mathbb{R}^{128}$ for the face interior region, and $\bm z_{additive} \in \mathbb{R}^{128}$ for the regions outside the face model (e.g., hair).
The latent code for the face $\bm z_{face}$ is converted by the face texture decoder into the face regions provided by the 3D model. 
The latent code for the rest of the head $\bm z_{additive}$ is processed into features for the rest of the face by the additive decoder.

 The coefficients for shape $\bm \alpha$ and expression $\bm \beta$ can be sampled from a distribution extracted from reference images via 3D model fitting~\cite{gerig2018morphablebfm2017}, or specified manually, which allows artistic control.

\myparagraph{Face Texture Decoder}
\label{sec:decoder_interior}
The face texture decoder is a modified ResNet-18~\cite{he2016deepresnet} where we expand the latent code $\bm z_{face}$ to spatial feature maps and stack them along the channel dimension. The feature maps are processed in a series of upsampling and residual blocks to the desired texture dimensions. The output is a pose and shape independent multi-dimensional feature map in UV space, which we call \emph{neural texture}.
We provide the detailed architecture in the supplementary material.

\myparagraph{UV Rendering and Texture Sampling}
\label{sec:sampling}
In order to project the texture onto the image plane, we use a 3D morphable face model with a UV parameterization~\cite{gerig2018morphablebfm2017}. Given model coefficients for shape $\bm \alpha$, expression $\bm \beta$, and a rotation matrix $\bm R$, we compute the posed mesh. We then project the UV parameterization to image space following the standard computer graphics pipeline
and use it to sample features from the neural face texture. The output of this step is a \emph{neural face feature image} $\bm F_{face}$.

\myparagraph{Additive Decoder}
\label{sec:decoder_exterior}
The face texture decoder yields a neural texture for the face region only. 
The additive decoder adds features for the regions missing in the face model, e.g., the hair or mouth interior. 
This is a very challenging task because the shape and appearance of the added regions should be consistent even for extreme head poses.
The additive decoder should therefore be invariant to pose-dependent features in the latent code. Please refer to Sec.~\ref{sec:augmentation} and the supplementary material for more details.

We condition the additive decoder on both the latent description of identity $\bm z_{additive}$ and the neural face feature image $\bm F_{face}$. The latent code $\bm z_{additive}$ is expanded to a spatial feature map (similar to the face texture decoder) and upscaled in a series of ResNet layers~\cite{he2016deepresnet}. In each block, we concatenate the rescaled face feature image as conditioning on geometry and pose.

The output of the additive texture decoder is an \emph{additive feature image} $\bm F_{additive}$ that is pixel-aligned given a pose, shape, expression, and identity.

\myparagraph{Feature2Image Network}
\label{sec:features2image}
The last step of the \oursname{} Generator pipeline is to convert the feature images $\bm F_{face}$ and $\bm F_{additive}$ to an RGB output image.
The Feature2Image network translates the stacked feature images into an RGB image and a foreground mask. Similar to~\cite{thies2019deferred,thies2020neural}, 
the Feature2Image network is a U-Net~\cite{ronneberger2015u}.

\subsection{Training}
\label{sec:training}
In contrast to existing methods that require strong supervision in the form of multi-view images or videos, we train only on unpaired monocular RGB images.

\myparagraph{Encoder}
\label{sec:encoder}
During training, we learn a latent space $\bm z \sim \mathcal{N}(\bm \mu_z, \bm \Sigma_z)$.
A ResNet-18~\cite{he2016deepresnet} encoder takes a foreground-masked RGB image and predicts the mean $\bm \mu_z \in \mathbb{R}^{256}$ and diagonal covariance $\bm \Sigma_z \in \mathbb{R}^{256\times 256}$, from which we sample a latent code $\bm z \in \mathbb{R}^{256}$ and process it further as described in Sec.~\ref{sec:generator}.

\myparagraph{Augmentation Scheme}
\label{sec:augmentation}
While the parametric face model allows for geometry-consistent synthesis for the face interior region, doing the same for the face exterior, where no 3D geometry is available, is much more challenging. A Variational Auto Encoder~\cite{kingma2014vae} trained by reconstruction would simply learn to copy such regions (e.g., hair) into the same spatial location even under different poses. 

To solve this problem, we employ an augmentation scheme to map our input image $\bm I$ to a transformed output image $A(\bm I)$. The mapping $A$ consists of random affine transforms: in-plane rotation, translation, scaling, and flipping. As a result, the additive decoder is guided to learn a one-to-many mapping---the same latent code $\bm z_{additive}$ must yield different additive feature images, which is determined by the pose and geometry from the face feature image. Please see \reffig{fig:pipeline} for a visual example and the supplementary for more details.

\myparagraph{Objective Function}\label{sec:obj_func}
Each training sample consists of a foreground-masked training image $\bm I$, its affine transformed version $A(\bm I)$, the ground-truth segmentation mask $\bm M$ belonging to $A(\bm I)$, and their corresponding reconstructions $\hat{\bm I}$ and $\hat{\bm M}$. We denote the spatial dimensions as $H$ and $W$.

For self-supervised reconstruction, we employ a photometric $\mathcal{L}_2$ loss term and a perceptual loss term $\mathcal{L}_{VGG}$:
\begin{equation}
\begin{split}
    \mathcal{L}_2 =& \|\hat{\bm I} - A(\bm I)\|^2_2,\\
    \mathcal{L}_{VGG} =& \sum_j v_j \|\phi_{VGG_j}(\hat{\bm I}) - \phi_{VGG_j}(A(\bm I))\|_1, 
    \label{eq:reconstruction}
\end{split}
\end{equation}
where the function $\phi_{VGG_j}(\cdot)$ extracts the $j$-th feature map from a pretrained VGG network~\cite{simonyan2014veryvgg,deng2009imagenet}, and $v_j$ are the weights per feature map (listed in the supplementary).

In order to learn correct foreground masks, we supervise with a cross entropy loss term $\mathcal{L}_{M}$:
\begin{equation}
\begin{split}
\label{eq:crossentropy}
    \mathcal{L}_M =& -\frac{1}{HW}\sum_{i}^{H}\sum_{j}^{W}{\bm M}_{ij}\text{log}\hat{\bm M}_{ij} \\
    &+ (1 - {\bm M}_{ij}) (1-\text{log}\hat{\bm M}_{ij}). \\
\end{split}
\end{equation}

We smooth the latent space with a Kullback-Leibler regularization term:
\begin{align}
\label{eq:kl}
    \mathcal{L}_{KL} = \mathcal{D}_{KL} \left( q(\bm z | \bm I) || p(\bm z) \right),
\end{align}
where $q(\bm z | \bm I)$ is the distribution predicted by the encoder and $p(\bm z)$ is a standard Gaussian distribution~\cite{kingma2014vae}.

To encourage realism, we employ a two-scale patch discriminator $D$ ~\cite{park2019semanticspade} with feature matching. The adversarial generator loss term is
\begin{equation}
\begin{split}
\label{eq:generatoradversarial}
 \mathcal{L}_{adv} =& (1 - D(\hat{\bm I}))^2 \\ 
 &+  \sum_j ||\phi_{D_j}({\hat {\bm I}}) - \phi_{D_j}(A( \bm I)) ||_1,
\end{split}
\end{equation}
where the function $\phi_{D_j}(\cdot)$ extracts the $j$-th feature map from the discriminator network.

The final losses for the generator and discriminator are:
\begin{equation}
\begin{split}
\label{eq:loss}
\mathcal{L}_\text{Generator} = & \lambda_2 \mathcal{L}_2 + \lambda_{VGG} \mathcal{L}_{VGG} + \lambda_M \mathcal{L}_M \\
    &+ \lambda_{KL} \mathcal{L}_{KL} + \lambda_{adv} \mathcal{L}_{adv}, \\
 \mathcal{L}_{\text{Discriminator}} = & \lambda_{adv} \frac{1}{2} \left[ D(\hat{\bm I})^2 +  \left(1 - D(A(\bm I))\right)^2 \right].
\end{split}
\end{equation}

We empirically choose $\lambda_2 = \lambda_M = \lambda_{adv} = 1$, $\lambda_{VGG} = 2$, and $ \lambda_{KL}=0.1$. For more training details and hyperparameters, please refer to the supplementary material.

\section{Experimental Setup}
\myparagraph{Data and Preprocessing}
We train our method on face images from the Flickr-Faces-HQ dataset (FFHQ)~\cite{karras2019style}. For training, we fit the Basel Face Model~\cite{gerig2018morphablebfm2017} offline. In nine cases, the model fitting fails ($<0.02\%$ of all images). We remove those images from the training set and end up with $59,991$ training and $10,000$ test samples, following the recommended splits. We visualize the removed images in the supplementary material. 

We aim to generate images with their corresponding foreground masks. 
To get pseudo-ground-truth, we train a state-of-the-art face segmentation network~\cite{chen2017deeplab,chen2017rethinkingdeeplabv3} on CelebAMask-HQ~\cite{lee2020maskgan} and predict the segmentation maps offline. Please refer to the supplementary material for details.

\myparagraph{Identity Consistency Metric}
\label{sec:metrics}
To evaluate identity consistency, we compute a similarity score from the embeddings of a state-of-the-art face recognition network~\cite{deng2019arcface}. For each related method, we render 3,000 identities with frontal head pose and compute their embeddings~\cite{deng2019arcface}. We then re-pose the same identities to various degrees and compute the cosine similarity between the normalized embeddings. As a reference for the reader, we provide similarities of a real-world multi-view dataset~\cite{zhang2020eth}. The real-world dataset contains faces with a slightly non-frontal pose (about $\pm$7\degree{}), hence, we use the average embedding of the two most frontal faces for the frontal pose.

\section{Results and Discussion}

\begin{figure}
    \centering
   \includegraphics[width=1\columnwidth]
                   {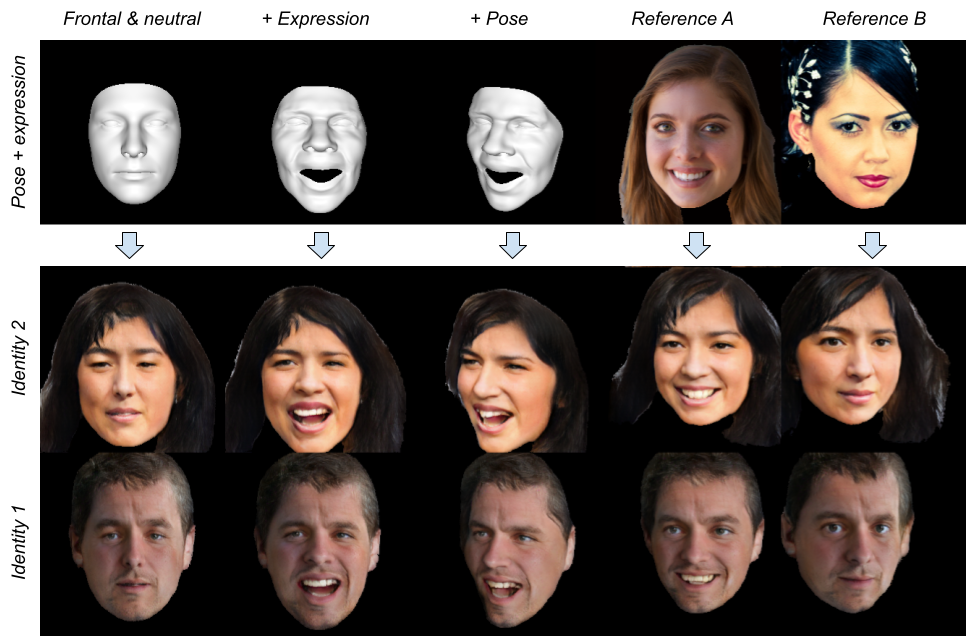}
    \caption{Rendering two identities under expression and pose control. Column 1 starts with a neutral pose and expression. Columns 2 and 3 change expression and pose via the graphical control unit (Fig.~\ref{fig:pipeline}). For columns 4 and 5, we render the face with expression and pose from real reference images. The top row shows the corresponding face meshes and reference images.
    }
    \label{fig:qualitative}
\end{figure}

\begin{figure*}
\centering
   \includegraphics[width=\textwidth]
                   {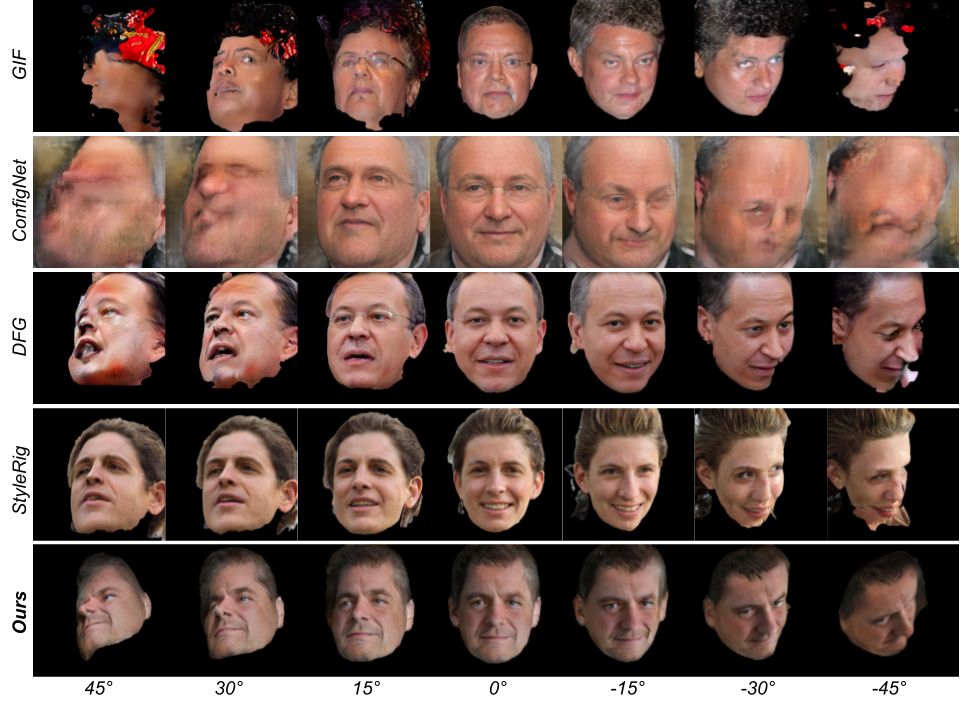}
    \caption{Comparison with related work. GIF~\cite{GIF2020},  ConfigNet~\cite{KowalskiECCV2020config}, and DiscoFaceGAN (DFG)~\cite{deng2020disentangleddiscofacegan} achieve impressive visual quality for re-posing faces, but only up to 15$^{\circ}$ from the frontal pose. StyleRig~\cite{tewari2020stylerig} renders photorealistic outputs, but is unable to render strong pose variations and instead falls back to a smaller pose variation value, for example as seen in the $+45^{\circ}$ case. Our technique is capable of synthesizing more \textit{extreme} poses while maintaining high identity consistency with the frontal image.
    \label{fig:comparison_pose}}
\end{figure*}

\begin{table*}
\small 
\begin{center}

\begin{tabular}{r ccccccc|ccccccc }
\hline
\multirow{2}{*}{\textbf{Method}} & \multicolumn{7}{c}{\textbf{Similarity yaw} $\leftrightarrow$} & \multicolumn{7}{c}{\textbf{Similarity pitch} $\updownarrow$} \\

  &  -45\degree & -30\degree & -15\degree & 0\degree & 15\degree & 30\degree & 45\degree & -45\degree & -30\degree & -15\degree & 0\degree & 15\degree & 30\degree & 45\degree \\
\hline

\cite{KowalskiECCV2020config} & 0.208 & 0.509 & 0.790 & - & 0.795 & 0.515 & 0.257 &  $\le 0$ & 0.014 & 0.459 & - & 0.476 & 0.095 & $\le 0$ \\
\cite{GIF2020}&  0.133 & 0.264 & 0.485 & - & 0.487 & 0.257 & 0.117 &   0.039 & 0.164 & 0.400 & -& 0.448 & 0.191 & 0.095 \\

\cite{deng2020disentangleddiscofacegan} &  0.530 & 0.690 & 0.866 & - & 0.863 & 0.675 & 0.521 &  
 0.270 & 0.461 & 0.781 &-& \textbf{0.826} & 0.581 & 0.388\\
\hline
\textbf{Ours} &  \textbf{0.568} & \textbf{0.729} & \textbf{0.874} & - & \textbf{0.873} & \textbf{0.732} & \textbf{0.585} &

  \textbf{0.416} & \textbf{0.611} & \textbf{0.821} & - & 0.817 & \textbf{0.611} & \textbf{0.420} \\
\hline
\textcolor{gray}{Ref} \cite{zhang2020eth} &\textcolor{gray}{0.855} & \textcolor{gray}{0.845} & \textcolor{gray}{0.726} & - & \textcolor{gray}{0.790} & \textcolor{gray}{0.773} & \textcolor{gray}{0.779}&
   \textcolor{gray}{0.719} & \textcolor{gray}{0.725} & \textcolor{gray}{0.753} &-& \textcolor{gray}{0.797} & \textcolor{gray}{0.805} & \textcolor{gray}{0.782} \\

\end{tabular}
\end{center}
\caption{Identity consistency for different head poses. We compare 3,000 frontal faces (0\degree) with randomly sampled expressions with their respective posed variants. The scores indicate the similarity calculated as the dot product between normalized embeddings from a state-of-the-art face recognition network~\cite{deng2019arcface} (higher is better). The bottom row (\emph{Ref}) is a reference to a real-world multi-view dataset~\cite{zhang2020eth}. For a visual comparison, please refer to Fig.~\ref{fig:comparison_pose}.\label{tbl:consistency}}
\end{table*}

\begin{figure}
    \centering
   \includegraphics[width=1\columnwidth]
                   {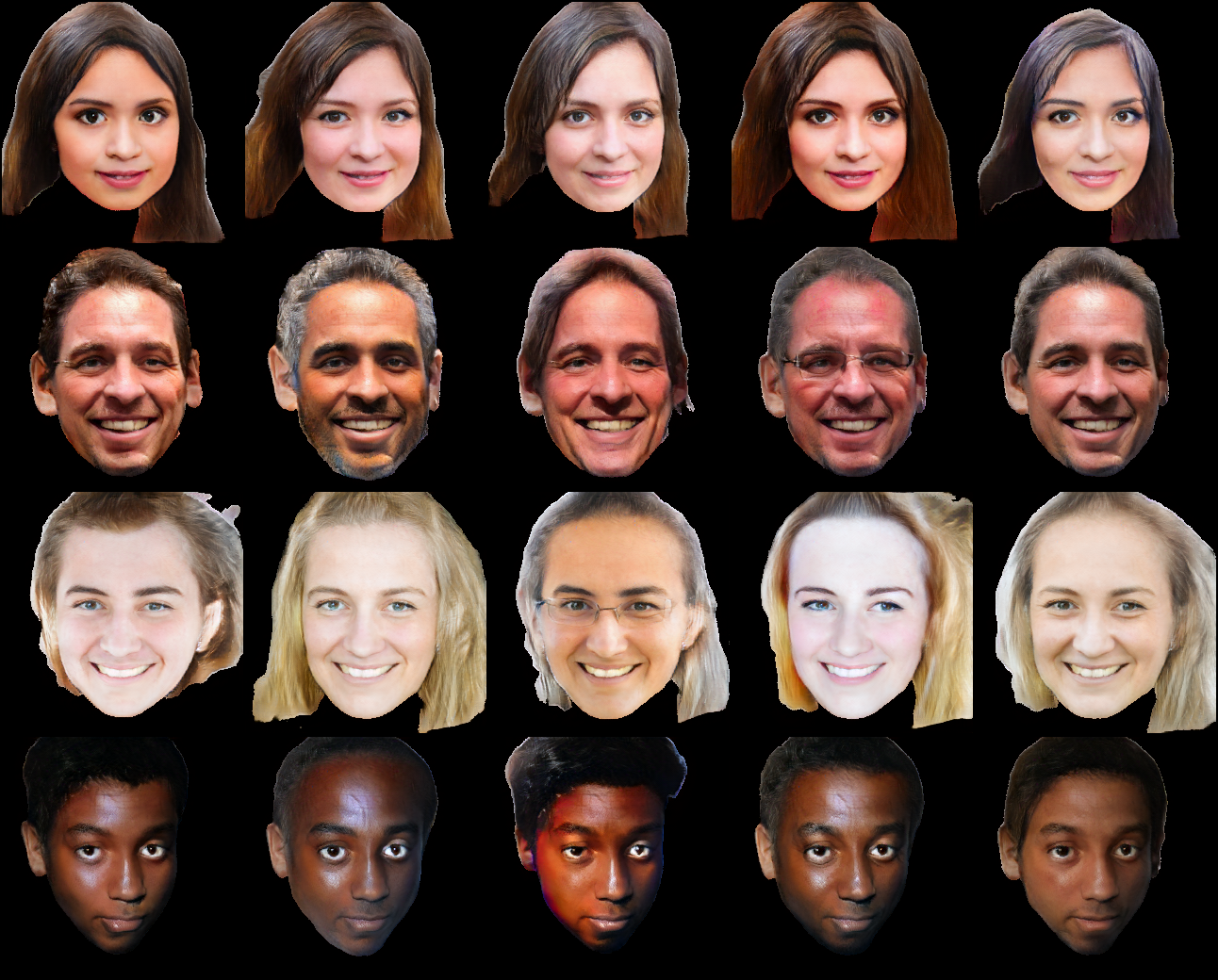}
    \caption{Sampling new identities. Each row samples from the learned latent distribution to generate variants of an identity. Note that the expression and pose are highly consistent.
    }
    \label{fig:sampling}
\end{figure}

\begin{table}
\small
\begin{center}
\begin{tabular}{lcc}
\hline
\textbf{Variant} &  \textbf{FID}$\downarrow$& \textbf{Consistency (yaw)} $\uparrow$\\
\hline
3-dim texture with $\mathcal{L}_{RGB}$&  54.27 & 0.712 $\pm$ 0.123 \\
3-dim texture w/o $\mathcal{L}_{RGB}$&  47.87 & 0.684 $\pm$ 0.132 \\
16-dim with $\mathcal{L}_{RGB}$ &  37.96 &  0.724 $\pm$ 0.119 \\
\hline
16-dim w/o $\mathcal{L}_{RGB}$ (\textbf{Ours}) &  \textbf{34.35} & \textbf{0.727} $\pm$ 0.121 \\

\end{tabular}
\end{center}
\caption{Ablation study. We compare photorealism (FID~\cite{heusel2017ttur,Seitzer2020FID}) and identity consistency for \emph{neural} vs. \emph{RGB} textures. A higher dimensional neural texture can yield photorealistic outputs, while also maintaining high consistency. We provide detailed ablation results and visual examples in the supplementary material.\label{tbl:ablation}}
\end{table}

This section discusses our results for controlled face synthesis. In Sec.~\ref{sec:qualitative}, we show qualitative results for rendering different geometries and poses (Fig.~\ref{fig:qualitative}) and sampling novel identities (Fig.~\ref{fig:sampling}). In Section~\ref{sec:results_rw}, we compare both qualitatively (Fig.~\ref{fig:comparison_pose}) and quantitatively (Tbl.~\ref{tbl:consistency}) with related work. In Sections~\ref{sec:results_userstudy} and~\ref{sec:ablation}, we conduct a user and an ablation study. In Section ~\ref{sec:limitations}, we discuss limitations and future work.

\subsection{Qualitative Results}
\label{sec:qualitative}
\myparagraph{Controlling Geometry and Pose}
\oursname{} can sample novel identities and produce consistent images for different geometries and poses. Fig.~\ref{fig:qualitative} shows sequential edits on two identities.
We start at a frontal pose and a neutral expression (column 1). We use the graphical control unit (Fig.~\ref{fig:pipeline}) to change expression and pose (columns 2 and 3). The top row shows the corresponding face mesh. It is also possible to extract the parameters of the graphical control unit from reference images (columns 4 and 5). Our method maintains high identity consistency across manipulations.

\myparagraph{Sampling and Identity Mixing}
\label{sec:identitymixing}
While previous works are limited to person-specific face textures~\cite{thies2019deferred,thies2020neural}, \oursname{} can generate textures for novel identities by sampling in a latent space (Fig.~\ref{fig:pipeline}). In Fig.~\ref{fig:sampling}, we sample new variants $\bm z_{j} \sim N(\bm \mu_{z_j}, \bm \Sigma_{z_j})$ for identities $j$ from the test set.
\oursname{} can also interpolate between two latent codes. We show such examples in the supplementary document and video.

\subsection{Identity Consistency}
\label{sec:results_rw}
A key benefit of using textured 3D geometries is that they allow highly consistent renderings even for extreme head poses. 
Our method leverages the strict mapping from texture to image space (Sec.~\ref{sec:sampling}). This facilitates the rendering of the identity-specific facial appearance for extreme poses, despite a dataset with mostly frontal faces. We visualize the head pose distribution of the training set and out-of-distribution samples in the supplementary.

We visually compare identity consistency in Fig.~\ref{fig:comparison_pose}. Related works~\cite{tewari2020stylerig,GIF2020,KowalskiECCV2020config,deng2020disentangleddiscofacegan} achieve highly consistent and photo-realistic results for frontal faces and poses up to ~30\degree{} (pitch) and ~15\degree{} (yaw). For more extreme poses, they tend to show severe artifacts~\cite{tewari2020stylerig,GIF2020,deng2020disentangleddiscofacegan} or blurred results~\cite{KowalskiECCV2020config}.

For StyleRig~\cite{tewari2020stylerig}, we exclusively show qualitative results because only a handful sample images were available to us. For the other methods~\cite{GIF2020,KowalskiECCV2020config,deng2020disentangleddiscofacegan} we generate 3000 samples and conduct a quantitative comparison by computing a similarity score using the identity consistency metric (Sec.~\ref{sec:metrics}). Tbl.~\ref{tbl:consistency} lists the resulting similarity scores (higher is better). Our method achieves the highest similarities, except for one of the evaluated poses. 

\subsection{User Study}
\label{sec:results_userstudy}
We conduct a perceptual user-study comparing our method with three state-of-the-art techniques for controlled face image synthesis~\cite{GIF2020,KowalskiECCV2020config,deng2020disentangleddiscofacegan} along two dimensions:

\begin{enumerate}[nolistsep]
\item The general quality of \textbf{photorealism} produced by the methods for images posed at random variations in the range of $[-45 ^{\circ}, 45 ^{\circ}]$ from the frontal pose. Participants answered the following question for 20 randomly chosen image pairs: \emph{Of the two images, which looks more like a real person?}
\item \textbf{Identity consistency} for triplets of images of the same identity synthesized at 3 different poses: frontal pose, -45\degree{} and 45\degree{} degrees along the yaw and pitch axis. Each user was shown 10 randomly chosen pairs of such triplet images generated by ours and the related works. We asked: \emph{Which set represents the same person more consistently?}

\end{enumerate}
The survey consisted of 128 participants. Our method is on par in terms of photorealism, and clearly outperforms competing baselines for identity consistency criteria. In the following, we report the user study results for pairwise comparisons of each related work against ours.

 For photorealism, 10\% of the participants voted in favour of ConfigNet~\cite{KowalskiECCV2020config}, 35\% preferred GIF~\cite{GIF2020} and 50\% chose DiscoFaceGAN~\cite{deng2020disentangleddiscofacegan}. 
 For identity consistency, 0\% of the participants preferred ConfigNet or GIF against \oursname{}; 8\% of participants preferred DiscoFaceGAN.
We provide results on other poses a random selection of example images from the survey in the supplementary document.

\subsection{Ablation Study}\label{sec:ablation}
We analyze the effect of neural textures in an ablation study. We simulate a traditional RGB texture by limiting the texture to 3 dimensions and imposing an additional constraint to make the texture resemble a classical RGB texture: $\mathcal{L}_{RGB} = \frac{1}{3}\sum_{c=1}^3||\bm F_c - A(\bm I_c)||^2_2$.

The variable $\bm F$ denotes the \emph{feature image} (Fig.~\ref{fig:pipeline}) and $A(\bm I)$ denotes the masked affine transformed training image (as described in Sec.~\ref{sec:obj_func}). The subscript $c = 1,...,3$ represents the three RGB channels.

We train four combinations: a) a 3-dimensional texture with $\mathcal{L}_{RGB}$, b) a 3-dimensional \emph{neural} texture without $\mathcal{L}_{RGB}$, c) a 16-dimensional texture with $\mathcal{L}_{RGB}$ and d) a 16-dimensional \emph{neural} texture without $\mathcal{L}_{RGB}$ (\textbf{ours}).

Table~\ref{tbl:ablation} compares photorealism (FID~\cite{heusel2017ttur,Seitzer2020FID}) and identity consistency over the head poses (as described in Sec.~\ref{sec:results_rw}). The consistency scores are the mean and the corresponding standard deviations over all poses. Please note that FID is computed on images masked to the foreground---the values are not directly comparable to related works that use backgrounds. 
Tree dimensional textures yield lower consistency and the generated images show artifacts---mostly visible in difficult regions, like eyes.
The results indicate that our network benefits from the higher expressiveness of neural textures. The FID score shows that high-dimensional textures improve realism. 
In the supplementary material, we provide additional results and further ablations.

\subsection{Limitations and Future Work}
\label{sec:limitations}
The proposed architecture allows going outside the training distribution. However, we observe a significant decrease in performance at very extreme poses beyond 60\degree. Furthermore, rigid objects inside the face get distorted by the perspective projection, e.g., when re-posing a face with glasses. We demonstrate examples for both cases in the supplementary material.

Possible extension to this work could generate complete images including backgrounds and torsos, and further disentangle the latent identity space.

\section{Conclusion}
We introduce~\oursname{}---a generative model of neural face textures. The \oursname{} framework affords sampling novel identities while controlling both pose and geometry.
Previous works excelled at either task individually; our framework generates novel identities \textit{and} renders them in a significantly larger range of controlled poses and expressions. Our method achieves this by learning to synthesize an arbitrary pose-independent neural texture from a latent code, sampled from a distribution that is learned in a fully self-supervised scheme from monocular face images. The neural texture is then rendered to an image with any desired pose and expression. Our method also consistently generates the challenging face exterior regions such as hair, ears, and mouth-interiors. We demonstrate the capabilities of our method through qualitative, quantitative, and perceptual analysis. We also identify the limitations and discuss the various possibilities emerging from this line of work. 
\\

\small
\noindent\textbf{Acknowledgments and Disclosure of Funding.}
We thank Xucong Zhang, Emre Aksan, Thomas Langerak, Xu Chen, Mohamad Shahbazi, Velko Vechev, Yue Li, and Arvind Somasundaram for their contributions. We also thank Ayush Tewari for the StyleRig visuals.
This project has received funding from the European Research Council (ERC) under the European Union’s Horizon 2020 research and innovation program grant agreement No 717054.

\begin{figure}[hb]
\centering
\includegraphics[width=0.5\columnwidth]{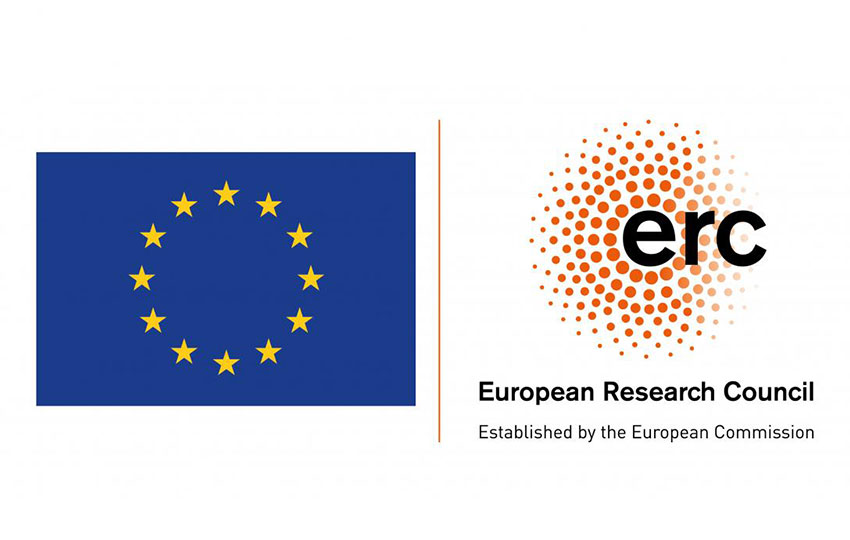}
\end{figure}

{\small
\bibliographystyle{ieee_fullname}
\bibliography{egbib}
}

\end{document}